\documentclass[lettersize,journal]{IEEEtran}
\usepackage{amsmath,amsfonts}
\usepackage{algorithmic}
\usepackage{algorithm}
\usepackage{array}
\usepackage[caption=false,font=normalsize,labelfont=sf,textfont=sf]{subfig}
\usepackage{textcomp}
\usepackage{stfloats}
\usepackage{url}
\usepackage{verbatim}
\usepackage{multirow}
\usepackage{graphicx}
\usepackage{cite}
\usepackage{booktabs}
\begin{document}

\title{MPT: Motion Prompt Tuning for Micro-Expression Recognition}

\author{
Jiateng~Liu$\dag$, ~\IEEEmembership{Student Member, ~IEEE},
Hengcan~Shi$\dag$, ~\IEEEmembership{Member, ~IEEE},
Feng~Chen,
Zhiwen~Shao,\\
Yaonan~Wang, 
Jianfei~Cai, ~\IEEEmembership{Fellow, ~IEEE},
Wenming~Zheng$^*$, ~\IEEEmembership{Senior Member, ~IEEE}
\thanks{
\dag~indicates the equal contribution.

Corresponding author: Wenming Zheng.


J. Liu, W. Zheng are with the Key Laboratory of Child Development and Learning Science of Ministry of Education, Southeast University, Nanjing 210096, China and also with the School of Biological Science and Medical Engineering, Southeast University, Nanjing 210096, China (e-mail: jiateng\_liu@seu.edu.cn; wenming\_zheng@seu.edu.cn).

H. Shi, Y. Wang are with the School of Artificial Intelligence and Robotics, Hunan University, (e-mail: shihengcan@hnu.edu.cn; yaonan@hnu.edu.cn).

F. Chen is with the Australian Institute for Machine Learning, the University of Adelaide, (e-mail: chenfeng1271@gmail.com).

Z. Shao is with the School of Computer Science and Technology, China University of Mining and Mine Digitization Engineering Research Center of the Ministry of Education, Xuzhou 221116, China, (e-mail: zhiwen\_shao@cumt.edu.cn).

J. Cai is with the Department of Data Science \& AI, Monash University, Melbourne 3800, Australia (e-mail: jianfei.cai@monash.edu).
}
}

\markboth{Journal of \LaTeX\ Class Files,~Vol.~14, No.~8, August~2021}%
{Shell \MakeLowercase{\textit{et al.}}: A Sample Article Using IEEEtran.cls for IEEE Journals}


\maketitle

\begin{abstract}
Micro-expression recognition (MER) is crucial in the affective computing field due to its wide application in medical diagnosis, lie detection, and criminal investigation. 
Despite its significance, obtaining micro-expression (ME) annotations is challenging due to the expertise required from psychological professionals. Consequently, ME datasets often suffer from a scarcity of training samples, severely constraining the learning of MER models. While current large pre-training models (LMs) offer general and discriminative representations, their direct application to MER is hindered by an inability to capture transitory and subtle facial movements-essential elements for effective MER. This paper introduces Motion Prompt Tuning (MPT) as a novel approach to adapting LMs for MER, representing a pioneering method for subtle motion prompt tuning.
  Particularly, we introduce motion prompt generation, including motion magnification and Gaussian tokenization, to extract subtle motions as prompts for LMs. Additionally, a group adapter is carefully designed and inserted into the LM to enhance it in the target MER domain, facilitating a more nuanced distinction of ME representation. 
  Furthermore, extensive experiments conducted on three widely used MER datasets demonstrate that our proposed MPT consistently surpasses state-of-the-art approaches and verifies its effectiveness.
\end{abstract}


\section{Introduction}

\begin{figure}[ht]
    \centering
    \includegraphics[width=0.47\textwidth]{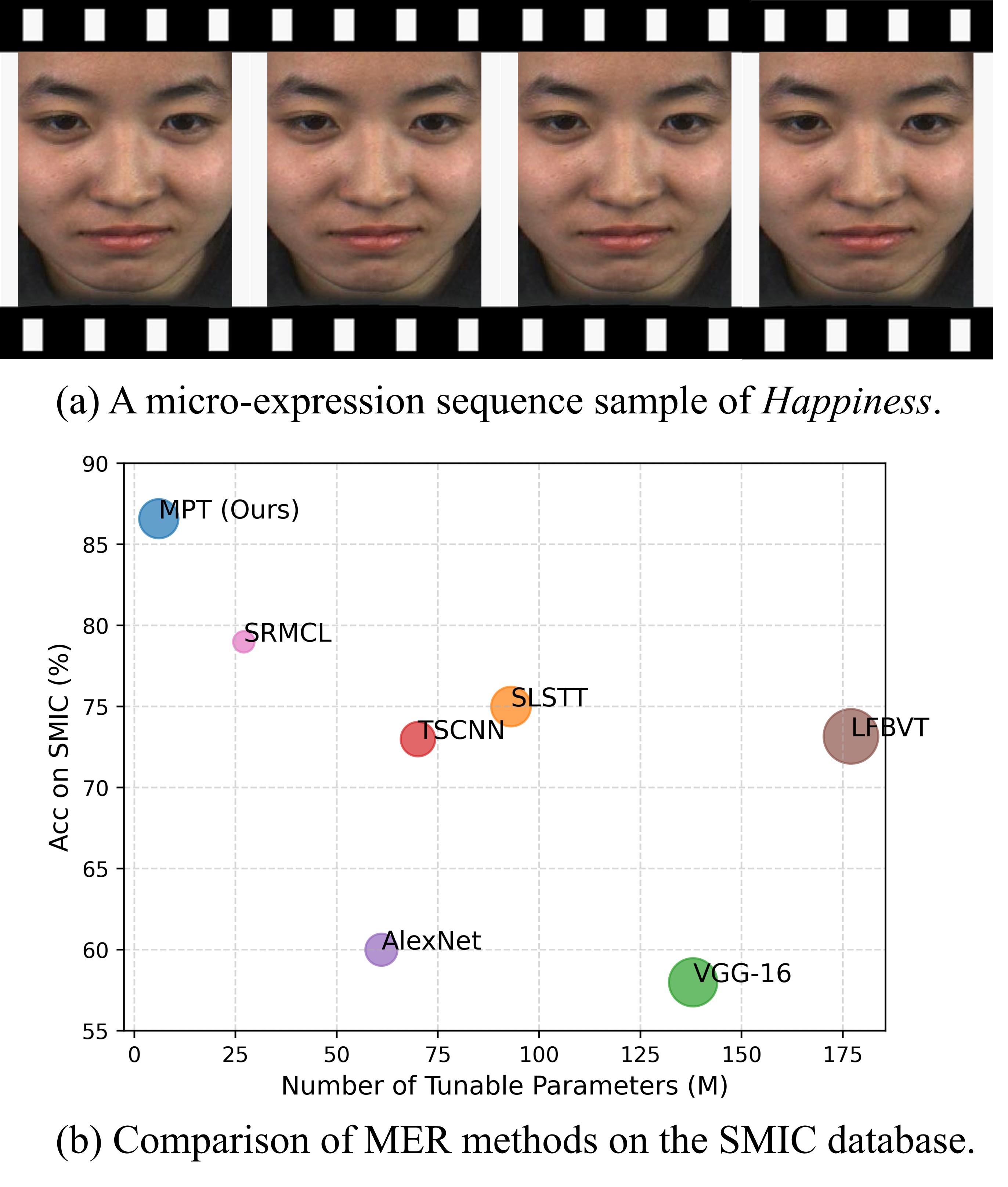}
    \caption{
      (a) An example of micro-expression of \textit{Happiness}. The key to MER is to capture subtle facial movements. 
    (b) Compared with previous methods, our proposed MPT achieves the best recognition accuracy (Acc) while containing fewer tunable parameters. Here we compare our MPT with SRMCL~\cite{bao2024boosting},  TSCNN~\cite{ThreeStream}, LFBVT~\cite{hong2022late}, SLSTT~\cite{SLSTT}, AlexNet~\cite{AlexNet}, and VGG-16~\cite{VGG-16}
    }
    \label{fig_introduction}
\end{figure}
Micro-expressions (MEs) are comprised of transitory and subtle facial muscle movements, as illustrated in Fig.~\ref{fig_introduction} (a). 
Haggard and Isaacs~\cite{haggard1966micromomentary} have shown that micro-expressions can unveil authentic human emotions that individuals may attempt to conceal, and accurate micro-expression recognition (MER) stands as a crucial step for many real-world applications, 
 such as lie detection, criminal investigation, and clinical diagnosis~\cite{LearnFromHierzong2018learning,zhou2024micromamba,liu2022cross}. Thus, MER has become a research hotspot in the field of affective computing, and many novel approaches have been proposed in recent years to cope with it.
Nevertheless, MER is still a very challenging task. 
This is because of the low intensity and short duration of MEs that may lead to subtle and irregular muscle movements in facial regions and temporal segments,
which also makes it extremely difficult to capture the comprehensive spatial-temporal representation of MEs. 
Consequently, it is desirable to learn a reliable representation of MEs to recognize the MEs more robustly.

Current MER methods can be generally divided into two types.
The first type is hand-crafted methods~\cite{LBP-TOP+TIM,LBP-SIP,LBP-TOPCASME,LBP-TOP-SAMM}, which manually extract facial texture and movements for recognition.
However, hand-crafted features are usually hard to capture complex patterns in facial movements, and thus limit their recognition ability. 
The second type is deep learning-based methods, such as CNN and Transformer-based solutions~\cite{ThreeStream,wei2023cmnet,SLSTT,GraphTCN}. 
They utilize learnable deep features to capture more comprehensive spatio-temporal relations in micro-expression sequences, generally outperforming hand-crafted methods. Nonetheless, deep-learning-based methods, especially Transformer-based ones, typically carry a mass of learnable parameters and require enormous training data. This brings in the challenge for MER since MER datasets usually only contain a small number of training samples, \textit{e.g.}, 247 and 164 samples in the CASME II and SMIC datasets, respectively, which are hard to train heavy deep MER models.

On the other hand, large pre-training models (LMs) have been widely developed in recent years, which are trained with millions of vision and/or language data to obtain general representations~\cite{vpt,clip,shi2022proposalclip,shi2023transformer,shi2024llmformer},
and have shown excellent performance on various downstream tasks like multi-modal tracking~\cite{maple}, action recognition~\cite{aim}, face generation~\cite{guo2023faceclip}, etc.
A straightforward way to address the few-sample issue is to leverage the general representations in LMs as prior knowledge and accelerate convergence~\cite{adapter,gao2024clip,guo2023faceclip}. 
However, LMs are usually trained with macroscopical data, while MER is required to recognize subtle and transitory facial movements as shown in Fig.~\ref{fig_introduction} (a).
Therefore, the key to leveraging LMs for MER is to incorporate subtle motion cues so that the LMs can capture more distinguishable ME representations. 
Motivated by the above discussion, in this paper, we propose an effective and efficient method termed motion prompt tuning (MPT) for adapting LMs for the MER task. MPT only fine-tunes a small number of parameters rather than the entire LM to preserve the knowledge of the LM, which can train the model more efficiently and reduce overfitting in small MER datasets. As shown in Fig.~\ref{fig_introduction} (b), MPT achieves an excellent performance using a smaller tunable parameter number than other MER methods.
Specifically, our MPT consists of a large pre-training vision model, a well-designed motion prompt generation module, and group adapters inserted in the LMs.  
The motion prompt generation module first magnifies and captures subtle facial motions in the ME sequence. Then, it models temporal expression flows by Gaussian tokenization to aggregate ME salient snapshots. Motion prompt tokens are generated, combined with facial appearance RGB tokens and a classification token, as the inputs of the LM. 
To transfer the knowledge learned by LM to the MER task, 
we further propose to adapt different types of tokens differently in the transformer layer by parameter-efficient group adapters. 
Specifically, the group adapter divides the classification token, motion prompts, and RGB tokens into separate groups in transformer layers and uses three learnable type-aware adapter operations to cope with them.
Compared with the previous common adapter structure~\cite{adapter}, the group adapter can aggregate different types of information in LMs more effectively.
Extensive experiments on three widely used benchmarks, CASME II, SAMM, and SMIC datasets, show that our MPT significantly outperforms state-of-the-art methods, demonstrating its effectiveness.

Our main contributions are summarized as follows: 
\begin{enumerate}
  \item To the best of our knowledge, this is the first work proposing to adapt LMs to the task of MER.
  \item For adapting LMs to MER, we propose MPT, a motion prompt generation and tuning method, which consists of motion magnification and Gaussian tokenization to extract motion prompts.
  %
  A simple yet effective group adapter is also proposed to adapt different types of representations into LMs efficiently.
  \item Extensive experiments on CASME II, SAMM, and SMIC demonstrate the effectiveness of our MPT. 
\end{enumerate}

The rest of the paper is organized as follows.
Section II briefly reviews the existing works on micro-expression recognition, large models, and parameter-efficient transfer learning.
Section III introduces the proposed motion prompt tuning (MPT) in detail.
Section IV reports extensive experimental results on several widely used datasets.
Finally, we draw a conclusion in Section V.

\section{Related Work}
\subsection{Micro-Expression Recognition} 
Current MER approaches can be mainly divided into two categories, 
 including hand-crafted and deep-learning-based methods.
On the one hand, Hand-crafted works manually design and extract features, such as local binary pattern (LBP) and its variants (LBP-TOP), histogram of oriented gradient (HOG), and histogram of optical flow (HOOF) to represent the spatial and temporal features of MEs~\cite{jiang2022seeking,liu2022cross}.
For example, 
Zong~\textit{et al.}~\cite{LearnFromHierzong2018learning} leverage LBP-TOP features and propose a hierarchical spatial grid division scheme and a kernelized group sparse learning (KGSL) model for MER. 
Liu~\textit{et al.}~\cite{MDMO} use sparse main directional mean optical flow (MDMO) to capture expression structures.
In~\cite{liu2022cross}, handcrafted features of MEs are used for conducting cross-database recognition for ME.
However, the hand-crafted features struggle to capture complex facial patterns to recognize various MEs in various scenes.
On the other hand, the deep-learning-based methods can learn more comprehensive expression representations to contain more accurate MER results~\cite{li2019bi,wei2022novel,zhou2024micromamba}. For example,
Lei \textit{et al.}~\cite{GraphTCN} extract the shape information of MEs and construct facial graph structures to model the intrinsic relations among important facial regions to help feature learning
Verma \textit{et al.} present several neural architecture search-based models to handle MER~\cite{verma2023efficient,verma2019learnet}.
Wei~\textit{et al.}~\cite{wei2023cmnet} use contrastive-learning-based distillation loss to encode explicit movement features and enforce a Wilcoxon rank sum test loss to calibrate the extracted intensity clues. These methods are based on CNNs, while many recent works~\cite{SLSTT,zhai2023feature,fan2023selfme} leverage Transformers to recognize MEs. 
Li~\textit{et al.}~\cite{li2020joint} employ multi-instance learning (MIL) to detect essential information on faces automatically and design a local maximum and global context joint learning to obtain a discriminative facial representation. 
In~\cite{chen2022block}, Chen et al. present a block division convolutional network (BDCNN) combined with an implicit semantic data augmentation loss to enhance ME feature learning.
Zhang \textit{et al.}~\cite{SLSTT} model short- and long-range relations to capture the local and global spatiotemporal patterns in MEs. They also utilize long-term optical flows to describe motions. FRL-DGT~\cite{zhai2023feature} uses a convolutional displacement generation module with self-supervised learning to extract the motion features between onset and apex frames. In SelfME~\cite{fan2023selfme}, Fan \textit{et al.} also use a self-supervised method combined with a symmetric contrastive vision transformer to extract facial motions in MEs.

Nevertheless, deep-learning-based methods require a mass of training data, while ME data is limited.
Unlike these approaches, we propose motion prompt tuning and a group adapter that are specifically designed to adapt LMs to MER, which only need to learn a few parameters and thus reduce the requirements of training data.

\begin{figure*}[t]
    \centering
    \includegraphics[height=0.42\textwidth,width=0.98\textwidth]{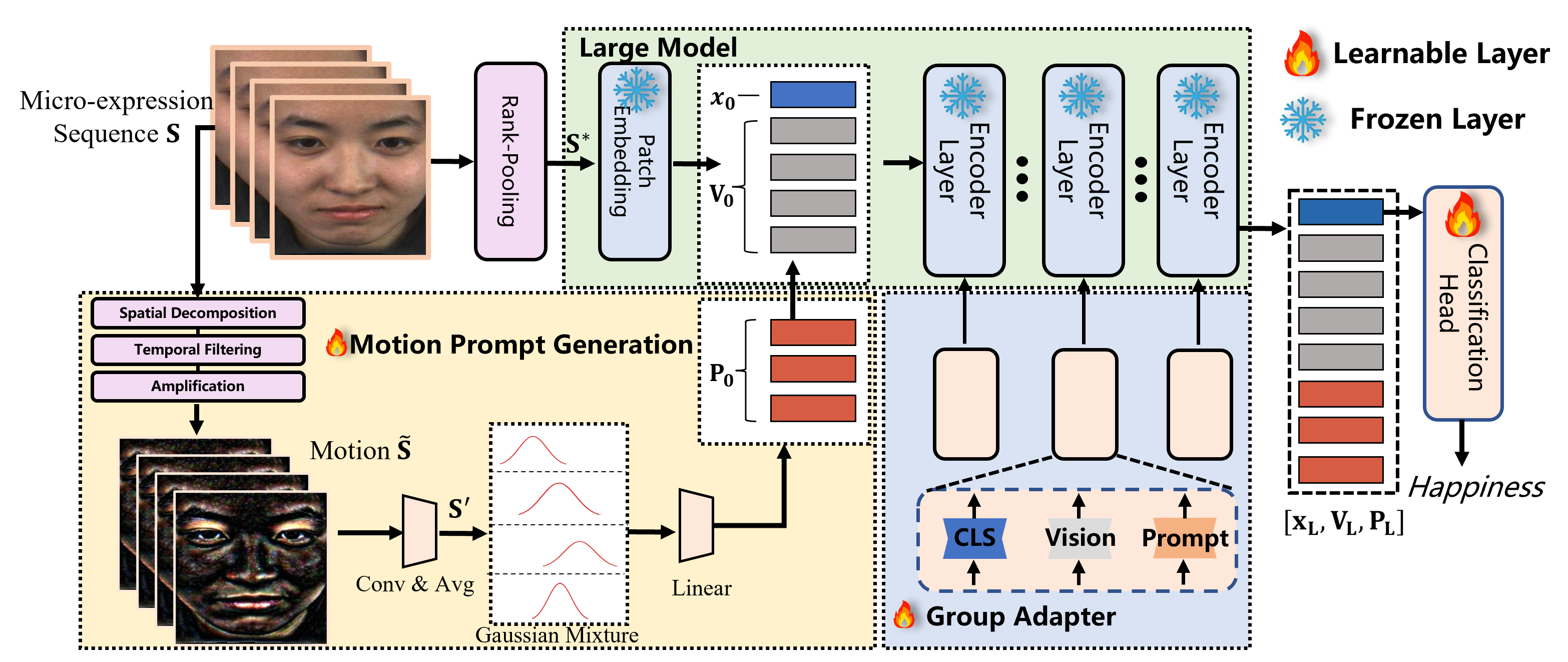}
    \caption{
      Pipeline of the proposed MPT. It consists of a motion prompt generation module, a large model backbone inserted with group adapters, and a learnable classification head. 
      Specifically, the input micro-expression sequence is processed by rank-pooling and patch-embedding operations
      to obtain class and vision tokens $\mathbf{x_0}$ and $\mathbf{V_{0}}$,
      and the motion prompt generation module generates the motion prompt tokens $\mathbf{P_0}$ at the same time.
      The concatenated tokens are then fed into a frozen $L$-layer transformer-based LM, in which
      each layer is inserted by a group adapter to help with the adaptation between different types of tokens.
      Finally, the classification head predicts the micro-expression type based on the output class token $\mathbf{x_L}$.
    }
    \label{fig_pipeline}
\end{figure*}

\subsection{Large Model and Parameter-Efficient Transfer Learning} 
In the rapidly evolving landscape of machine learning, large models (LMs)~\cite{clip,bert,shi2024llmformer} have garnered significant attention, emerging as a cornerstone in various tasks like natural language processing, image processing, and multimodal. 
These models are often based on transformer architectures like ViT~\cite{vit} and trained on millions of data. They have shown great performance in various downstream tasks, including image classification~\cite{vit}, object detection~\cite{carion2020end,li2021benchmarking}, facial expression analysis~\cite{guo2023faceclip}, and segmentation~\cite{shi2022proposalclip}.
Given their superior performance and much larger parameter scale, the question of efficiently adapting such models to the data-limited task becomes crucial. Existing efficient methods for adapting LMs into downstream tasks fall into two categories. The first type is prompt tuning, which refers to designing instructions and prepending them to the input of LMs.
Recently, prompt techniques~\cite{vpt,dpt,shi2022proposalclip} have attracted increasing research attention, which only fine-tune a few additional tokens in the input of LMs. 
VPT~\cite{vpt} adds a few randomly initialized vision tokens and achieves better performance than fully fine-tuned methods. ViPT~\cite{zhu2023visual} learns modal-related prompt tokens to adapt LMs to downstream tracking tasks. 
STPN~\cite{sun2023spatio} generates dynamic visual prompts gathering spatial and temporal information for robust video representation. These methods focus on capturing semantics and appearances as prompts, while we propose to generate motion prompts that can help to represent ME.
Another type of parameter-efficient transfer learning method is adapter tuning~\cite{adapter,chen2022vision,sung2022vl}, which adds a small number of learnable parameters to fixed LMs for adaptation.
Sung~\textit{et al.}~\cite{sung2022vl} leverage residual connections to add several linear layers to pretrained LMs, and thus they can fine-tune the additional linear layers while fixing LMs.
Pan~\textit{et al.}~\cite{St-adapter} and Yang~\textit{et al.}~\cite{aim} propose spatio-temporal adapters for image-video transfer learning tasks. They introduce frameworks that jointly spatial adaptation and temporal adaptation to gradually improve the spatiotemporal reasoning capability of the image model.
CLIP-Adapter~\cite{gao2024clip} combines both visual and textual feature adapters to adapt the pretrained CLIP~\cite{clip} model to target tasks.

Nevertheless, existing fine-tuning methods can not adapt to the MER domain well since they rely on naive prompts or simple adapter structures, which can not capture the subtle motions in MEs.
To handle this issue, we combine motion prompt tuning with unique group adapters, which are designed based on the characteristics of MEs for further improving the adaptation of LMs for ME.


\section{Method}
We propose MPT to effectively adapt LMs for the MER task. Instead of training the whole LM from scratch, MPT only fine-tunes a few parameters while freezing the parameters of the LM at the same time, which (1) reserves the pre-trained knowledge in the LMs, (2) trains the model more efficiently, and (3) reduces overfitting in small-scale ME databases. In this section, we will introduce the proposed MPT in detail.


\subsection{Overview}
As shown in Fig~\ref{fig_pipeline}, our MPT consists of three major components, \textit{i.e.}, (1) a Large Model (e.g., a ViT-16~\cite{vit}), (2) the Motion Prompt Generation, and (3) the Group Adapter. 
Specifically, consider a micro-expression sequence $\mathbf{S} =\{I_0, I_1, ... I_T\} \in \mathbb{R}^{T \times H \times W \times C}$, where $T$ is the number of frames, $I_t$ is the $t$-th frame in the sequence, $H$ and $W$ are the height and width of each frame, and $C$ is the number of channels.
Firstly, we follow ~\cite{verma2019learnet} to aggregate the sequence $\mathbf{S}$ into a single RGB image $\mathbf{S}^{*} \in \mathbb{R}^{H \times W \times C} $ by rank-pooling~\cite{dynamicnetwork1}. 
The basic idea of the rank-pooling approach is to aggregate the facial texture and appearance of a video sequence into a single RGB image by learning a ranking function to capture the temporal ordering of the sequence. 
By focusing on the order of features, rank-pooling decreases noise and variations in ME, and so it can effectively capture the temporal dynamics of MEs. 
Here we utilize the fast version for calculation~\cite{dynamicnetwork1}:
\begin{equation}
\begin{gathered}
\mathbf{S^*}=\rho\left(I_1, I_2, \ldots, I_t, \psi\right)=\sum_{t=1}^T \alpha_t \psi\left(I_t\right),\\
\alpha_t=2 t-T-1,
\end{gathered}
\end{equation}
where $\psi(I_t)$ denotes the feature representation of $I_t$.

Consequently, for a pre-trained $L$-layer Vision Transformer model~\cite{vit}, the calculated RGB image $\mathbf{S}^{*}\in \mathbb{R}^{H \times W \times C}$ is divided into fixed-sized patches,
each of which is embedded into a $D$-dimensional vector with positional embedding:
\begin{equation}
 \mathbf{V}_0^j=\operatorname{Embed}\left(\mathbf{S}^{*}\right), \quad \mathbf{V}_0^j \in \mathbb{R}^D, j=1,2, \ldots N_v,
\end{equation}
where $\operatorname{Embed}(\cdot)$ is the patch-embedding operation in vision transformer, $N_{v}$ is the number of visual tokens and $D$ is the dimension of each token.
In addition, a class token $\mathbf{x_0} \in \mathbb{R}^D$ is concatenated to the patch tokens sequence.

At the same time, motion prompts $\mathbf{P}_{0} \in \mathbb{R}^{N_{p} \times D}$ are generated by our motion prompt generation module to help the pre-trained LM to recognize subtle motion movements, where $N_{p}$ represents the number of motion tokens. The entire input of the LM is a concatenation of vision tokens $\mathbf{V}_{0}$, motion prompts $\mathbf{P}_{0}$ and an additional $[CLS]$ token $\mathbf{x}_{0} \in \mathbb{R}^{D}$. \textit{i.e.},
\begin{equation}
    \mathbf{I}=[\mathbf{x}_{0}, \mathbf{V}_{0}, \mathbf{P}_{0}],
\end{equation}
where $[\cdot,\cdot,\cdot]$ is the concatenation operation and $\mathbf{I} \in \mathbb{R}^{(N_{p}+N_{v}+1) \times D}$ is fed into the LM as input.

The LM comprehensively analyzes vision, motion, as well as class information and updates these tokens. As in the operation in ViT~\cite{vit}, we pass the updated $[CLS]$ token into a classification head to generate the final recognition result. However, LMs are usually trained on macroscopically RGB data, while inputs for MER are a facial RGB image and motion prompts. To alleviate the domain gaps, we propose a group adapter to further adapt LM representations to the MER domain. Different from the vanilla adapter~\cite{adapter}, our group adapter individually deals with each type of token to better adapt each category of data and reduce computational costs.
During our training, the large model is fixed, while only motion prompt generation, group adapter, as well as classification head are updated. Below, we introduce the details of our motion prompt generation and group adapter.

\subsection{Motion Prompt Generation}
The proposed motion prompt generation contains two steps, \textit{i.e.}, motion magnification to capture subtle movements for MER, as well as temporal Gaussian tokenization to further aggregate temporal movements and embed them into motion prompt tokens. 
\\
\textbf{Motion Magnification and Extraction.} In terms of~\cite{li2018canapex,li2020joint,EVM}, the rate of motion changes in an ME sequence can be represented by the special frequency components. To this end, we first magnify and extract the motion cues using an Eulerian approach, in which we analyze video variations at fixed pixel locations over time, and capture subtle changes in the frequency domain.
Specifically, let $I(x,y,t)$ represent pixel intensity at time $t$ and spatial coordinates $(x,y)$, we first decompose each frame $I_t$ in the ME sequence into $K$-level spatial frequency bands by using Laplacian pyramids $Lp(\cdot)$:
\begin{equation}
    \{L_1, ..., L_K\}=Lp(I_t).
\end{equation}
Then we apply temporal filtering for each spatial frequency level $k$:
\begin{equation}
    L_{k}^{'}=L_k(x,y,t)*H(t),
\end{equation}
where $H(t)$ denotes the temporal filter.
The filtered signals are magnified by a factor $\beta$:
\begin{equation}
    L_{k}^{''}(x,y,t) = \beta \cdot L_k^{'}(x,y,t),
\end{equation}
Finally, the magnified signals are reconstructed to obtain the motion sequence $\tilde{\mathbf{S}}$:
\begin{equation}
    \tilde{\mathbf{S}} = Rec({L_{1}^{''}, L_{2}^{''}, ..., L_{K}^{''}}),
\end{equation}
where $Rec(\cdot)$ denotes the Laplacian reconstruction function to form the final magnified signal.
As shown in Fig.\ref{fig_pipeline}, the motion magnification and extraction make it possible to observe the invisible changes over the ME sequence by the captured temporal motion sequence $\tilde{\mathbf{S}} \in \mathbb{R}^{T \times H \times W \times C}$. 
\\
\textbf{Temporal Gaussian Tokenization.}
We then integrate the captured temporal motions $\tilde{\mathbf{S}}$ and generate motion prompt tokens $\mathbf{P}_{0}$. Since ME sequences have several salient snapshots, like onset and apex frames, which contain more ME motion information~\cite{li2020joint,OFF-ApexNet} and play more important roles in representing ME, the temporal Gaussian tokenization is proposed to achieve the goal that adaptively aggregating these significant ME motion patterns over temporal dynamics.  
In particular, we generate $N_p$ motion prompt tokens, and for each token, we use a temporal Gaussian kernel to integrate motions. Therefore, there are $N_p$ Gaussian kernels.

Firstly, a $3 \times 3$ convolution layer $Conv(\cdot)$ and an average pooling $Avg(\cdot)$ are used to embed $\tilde{\mathbf{S}}$ and fuse spatial information:
\begin{equation}
    \mathbf{S}' = Avg(Conv(\tilde{\mathbf{S}})),
\end{equation}
where $\mathbf{S}' \in \mathbb{R}^{T \times D}$ is the embeded feature map.

Subsequently, two linear layers $\mathbf{W^{\mu}_{i}}$ and $\mathbf{W^{\sigma}_{i}}$ are used to generate the means and variances for $i$-th temporal Gaussian kernels:
\begin{align}
      \mathbf{\mu_{i}}      & = \mathbf{S}'\mathbf{W^{\mu}_{i}}, \\
      \mathbf{\sigma_{i}} & = \mathbf{S}'\mathbf{W^{\sigma}_{i}},
\end{align}
where $\mu_{i} \in \mathbb{R}^{D}$ and $\sigma_{i}^2 \in \mathbb{R}^{D}$ denote a pair of mean and variance of Gaussian distributions.
Then, a motion prompt token can be generated as:
\begin{align}
     \mathbf{p_{i}} &= G\left(\mathbf{S}'; \mu_{i}, \sigma_{i}^2 \right)\mathbf{W^{p}}\\
      &= \frac{1}{\sqrt{2 \pi \sigma_{i}^2}} \exp \left(-\frac{1}{2} \frac{(\mathbf{S}'-\mu_{i})^2}{\sigma_{i}^2}\right)\mathbf{W^{p}}.
\end{align}
where $\mathbf{W^{p}}\in \mathbb{R}^{D}$ is a learnable linear layer.
In the end, $N_p$ motion tokens are generated as the motion prompts of LMs.


\begin{figure*}[ht]
    \centering
    \includegraphics[height=0.35\textwidth, width=1.0\textwidth]{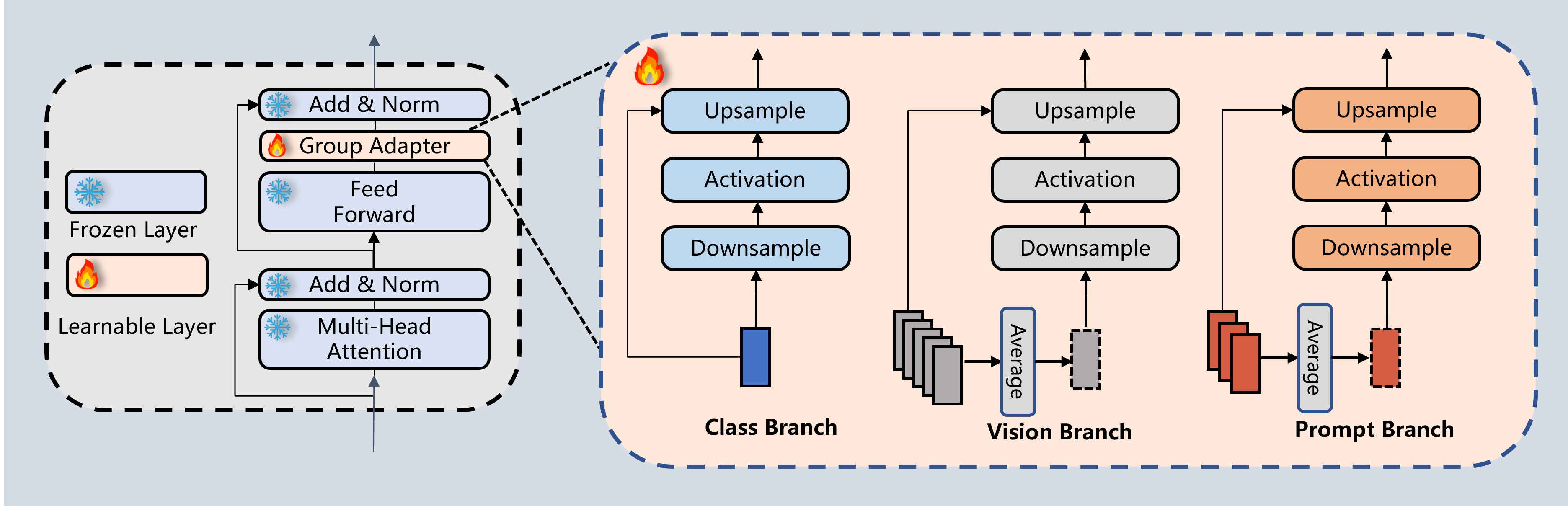}
    \caption{
      Overview of the group adapter. 
      It contains three branches, including the class, prompt, and vision branches to separate the transfer learning in the transformer layer for a more effective adaptation.
    }
    \label{fig_adapter}
\end{figure*}
\subsection{Group Adapter}
We input the concatenation of vision tokens $\mathbf{V}_{0}$, motion prompts $\mathbf{P}_{0}$ as well as the $[CLS]$ token $\mathbf{x}_{0}$ into the LM. The LM containing $L$ layers learns and updates representations as
\begin{equation}
    [\mathbf{x}_{l}, \mathbf{V}_{l}, \mathbf{P}_{l}] = Trans_{l}([\mathbf{x}_{l-1}, \mathbf{V}_{l-1}, \mathbf{P}_{l-1}]),
\end{equation}
where $Trans_{l}(\cdot)$ represents the $l$-th Transformer layer, and $l=1,..., L$. $[\mathbf{x}_{l-1}, \mathbf{V}_{l-1}, \mathbf{P}_{l-1}]$ are tokens from the previous block, and $[\mathbf{x}_{l}, \mathbf{V}_{l}, \mathbf{P}_{l}]$ are the ones updated by the $l$-th layer.

To transfer the knowledge learned by LM to the MER task, a straightforward insight is to inject domain-specific adapters to learn the information from different domains effectively~\cite{adapter}.
Adapters are learnable bottlenecks consisting of a downsampling layer, an upsampling layer, and an activation function to process different types of tokens in the transformer layer in a parameter-efficient way. 
However, the primitive adapter~\cite{adapter} processes all the tokens together, which cannot explore the complex modality-specific nuances. 
To this end, we divide the classification token, motion prompts, and RGB tokens into separate groups in transformer layers and use three learnable type-aware adapter operations to cope with them.

To this end, we propose a group adapter to reduce the gaps between MER and the LM.  
Specifically, as shown in Fig~\ref{fig_adapter}, group adapter is inserted after the feed-forward layer 
 to maximize the benefits of the transformer block's processing.
Furthermore, to facilitate efficient cross-modal learning, the group adapter is divided into three branches: a class branch, a motion branch, and a visual branch.
Each branch consists of a downsample linear layer, an activation function, and an upsample linear layer.
Formally, given the input feature matrix at the $l$-th layer, the feature adaptation process can be written as:
\begin{equation}
    \text {Adapt}^{\text{x}}_{l}(\mathbf{x}_{l})=\mathbf{x}_{l}+f\left(\mathbf{\hat{x}}_{l}\mathbf{W}^\text{xd}_{l}\right) \mathbf{W}^\text{xu}_{l},
\end{equation}
\begin{equation}
    \text {Adapt}^{\text{v}}_{l}(\mathbf{V}_{l})=\mathbf{V}_{l}+f\left(\mathbf{\hat{V}_{l}}\mathbf{W}^\text{Vd}_{l}\right) \mathbf{W}^\text{Vu}_{l},
\end{equation}
\begin{equation}
    \text {Adapt}^{\text{p}}_{l}(\mathbf{P}_{l})=\mathbf{P}_{l}+f\left(\mathbf{\hat{P}_{l}}\mathbf{W}^\text{Pd}_{l}\right) \mathbf{W}^\text{Pu}_{l},
\end{equation}
where $f(\cdot)$ denotes the GELU~\cite{gelu} activation function, 
 $\mathbf{W}^\text{xd}_{l}$,
 $\mathbf{W}^\text{Vd}_{l}$,
 $\mathbf{W}^\text{Pd}_{l}$~$\in \mathbb{R}^{D \times \frac {D}{\eta}}$ 
 refer to the downsample linear layers, and
 $\mathbf{W}^\text{xu}_{l}$,
 $\mathbf{W}^\text{Vu}_{l}$,
 $\mathbf{W}^\text{Pu}_{l}$~$\in \mathbb{R}^{{ \frac {D}{\eta}} \times D}$
 refer to the upsample linear layers, $\eta$ denotes the reduction factor.
 $\mathbf{\hat{x}}_{l}$, $\mathbf{\hat{V}}_{l}$ and $\mathbf{\hat{P}}_{l}$ denote the average of the class, 
 vision, and motion tokens in the $l$-th transformer layer, respectively.

\subsection{Classification Head and Loss Function}
The final $[CLS]$ token $\mathbf{x}_{L}$ is fed into a classification head containing a linear layer and a Softmax normalization. The classification head predicts the MER results in $\mathbf{y}^p$. The proposed method is optimized as follows:
\begin{equation}
    \mathcal{L}=CE\left(\mathbf{y}^g_{i}, \mathbf{y}^p_{i}\right),
\end{equation}
where $CE(\cdot)$ denotes the cross-entropy function, $\mathbf{y}^g_{i}$ and $\mathbf{y}^p_{i}$ are the ground truth and predicted result of $i$-th sample.
During training, only the parameters of motion prompt generation, group adapter, and classification head are optimized while all the other components are frozen. 

\begin{table}[ht]
    \centering
    \caption{Comparison results on the CASME II database.}
    \scalebox{1.0}{
    \begin{tabular}{@{}c|c|cc@{}}
    \toprule
    Methods                       & Year      & Acc(\%)              & F1-score (\%)           \\ \midrule
    LBP-TOP~\cite{LBP-TOPCASME}   & 2016      & 51.00                & 47.00               \\
    CNN+LSTM~\cite{CNN+LSTM}      & 2016      & 60.98                & -                   \\
    DSSN~\cite{DSSN}              & 2019      & 70.78                & 72.97               \\
    TSCNN~\cite{ThreeStream}      & 2019      & 80.97                & 80.70               \\
    KD~\cite{sun2020dynamic}      & 2020      & 72.61                & 67.00               \\
    Graph-TCN~\cite{GraphTCN}     & 2020      & 73.98                & 72.46               \\
    SLSTT~\cite{SLSTT}            & 2022      & 75.81                & 75.30               \\
    CMNET~\cite{wei2023cmnet}            & 2023      & 78.05                & 73.99               \\
    $\mu$-BERT~\cite{nguyen2023micron}& 2023  & -                    & 85.53               \\
    PLMaM-Net~\cite{wang2024progressively}& 2024 & 82.52             & 79.98               \\
    SRMCL~\cite{bao2024boosting}  & 2024      & 83.20                & 82.86               \\
    FFDIN~\cite{li2024structure}  & 2024      & 85.77       & 83.09     \\ 
    LTR3O~\cite{zhu2025learning}  & 2025      & 81.78       & 79.05     \\ \midrule
    \textbf{MPT (Ours)}           & 2025      & \textbf{87.85}       & \textbf{86.04}      \\ \bottomrule
    \end{tabular}}
    \label{tb_casme}
\end{table}

\begin{table}[ht]
    \centering
    \caption{Comparison results on the SAMM database.}
    \scalebox{1.0}{
    \begin{tabular}{@{}c|c|cc@{}}
    \toprule
    Methods                                & Year & Acc(\%)     & F1-score (\%)   \\ \midrule
    LBP-TOP~\cite{LBP-TOP-SAMM}            & 2016 & 34.56       & 28.92     \\
    LBP-SIP~\cite{LBP-SIP}                 & 2016 & 36.03       & 31.33     \\
    DSSN~\cite{DSSN}                       & 2019 & 57.35       & 46.44     \\
    TSCNN~\cite{ThreeStream}               & 2019 & 63.53       & 60.65     \\
    AUGCN~\cite{lei2021micro}              & 2021 & 74.26       & 70.45     \\
    Graph-TCN~\cite{GraphTCN}              & 2020 & 75.00       & 69.85     \\
    SLSTT~\cite{SLSTT}                     & 2022 & 72.39       & 64.00     \\
    CMNET~\cite{wei2023cmnet}                     & 2023 & 78.68       & 77.19     \\
    $\mu$-BERT~\cite{nguyen2023micron}     & 2023 & -           & 83.86    \\
    PLMaM-Net~\cite{wang2024progressively} & 2024 & 76.47       & 69.61     \\
    SRMCL~\cite{bao2024boosting}           & 2024 & 74.63       & 65.99     \\
    FFDIN~\cite{li2024structure}           & 2024 & 79.33       & 79.27     \\ 
    LTR3O~\cite{zhu2025learning}           & 2025 & 80.15       & 75.74     \\ \midrule
    \textbf{MPT (Ours)}                    & 2025 & \textbf{87.50}   & \textbf{85.04}         \\ \bottomrule
     \end{tabular}}
    \label{tb_samm}
\end{table}

\begin{table}[ht]
    \centering
    \caption{Comparison results on the SMIC database.}
    \scalebox{1.0}{
    \begin{tabular}{@{}c|c|cc@{}}
    \toprule
    Methods                           & Year & Acc(\%)                       & F1-score (\%)       \\ \midrule
    LBP-TOP~\cite{LBP-TOP+TIM}        & 2014 & 53.56                         & -                   \\
    STCLQP~\cite{STCLQP}              & 2017 & 64.02                         & 63.81               \\
    DSSN~\cite{DSSN}                  & 2019 & 63.41                         & 64.62               \\
    TSCNN~\cite{ThreeStream}          & 2019 & 72.74                         & 72.36               \\
    KD~\cite{sun2020dynamic}          & 2020 & 76.10                         & 71.00               \\
    SLSTT~\cite{SLSTT}                & 2022 & 75.00                         & 74.00               \\
    CMNET~\cite{wei2023cmnet}                & 2023 & 79.27                         & 80.11               \\
    $\mu$-BERT~\cite{nguyen2023micron}& 2023 & -                           & 85.50               \\
    PLMaM-Net~\cite{wang2024progressively}& 2024 & 82.32                     & 82.21               \\
    SRMCL~\cite{bao2024boosting}      & 2024 & 78.98                         & 78.87               \\
    FFDIN~\cite{li2024structure}      & 2024 & 79.07                         & 73.84               \\ 
    LTR3O~\cite{zhu2025learning}      & 2025 & 81.10       & 81.32     \\ \midrule
    \textbf{MPT (Ours)}               & 2025 & \textbf{86.59}                & \textbf{86.01}      \\
    \bottomrule
    \end{tabular}}
    \label{tb_smic}
\end{table}

\section{Experiments}
\subsection{Experimental Settings} 
Three widely used databases were utilized during the experiments, 
 including CASME II~\cite{CASMEII}, SAMM~\cite{SAMM}, and SMIC~\cite{SMIC}.
\textbf{The CASME II database} contains 247 micro-expression samples recorded from 27 subjects.
These ME samples are recorded in 200 FPS and 210$\times$200 resolution for facial regions, 
 and they are categorized into five emotion classes with labeled AUs, 
 including \emph{Happiness} (32 samples), \emph{Surprise} (25 samples), \emph{Disgust} (64 samples), \emph{Repression} (27 samples), and \emph{Others} (99 samples).
\textbf{The SAMM database }is built by Davison \textit{et al.} from  Manchester City University with 29 subjects from various races which includes 29 subjects from various races.
There are 159 ME samples with 8 emotion classes with labels AUs, which are recorded at 200 FPS and 230$\times$210 resolution for facial regions. 
Specifically, following the works in~\cite{ThreeStream,GraphTCN,bao2024boosting}, only samples more than 10 are selected in our experiments, 
 \textit{i.e.}, \emph{Anger} (57 samples), \emph{Contempt} (12 samples), \emph{Happiness} (26 samples), \emph{Surprise} (15 samples) and \emph{Others} (26 samples).
\textbf{The SMIC database} is collected by Li \textit{et al.} from the University of Oulu, Finland. 
It contains 164 ME samples for 16 subjects, each recorded at 100 FPS and 160$\times$140 resolution. 
Three classes of emotion are collected, \textit{i.e.}, \textit{Positive (51 samples), Negative (70 samples), and Surprise (43 samples).}
\\
\textbf{Implementation Details and Evaluation Metrics.}
The proposed MPT model is trained in an end-to-end manner, using PyTorch on an NVIDIA GeForce RTX 4090 GPU. Our motion prompt generation and group adapter modules are designed to integrate with any LMs. In this study, we utilize the pretrained Vision Transformer (ViT-B/16)~\cite{vit} as the backbone. All cropped ME samples are resized to a 224 $\times$ 224 size. The Adam optimizer~\cite{zhang2018improved} is employed for training the MPT with a batch size of 16, and the initial learning rate is set to 3e-4. The magnification factor 
$\beta$ is empirically set to 20, and the reduction factor $\eta$ is set to 8. 
Each ME sequence is standardized to a length of 16 frames at regular intervals, with temporal interpolation applied to sequences shorter than 16 frames. 
The Butterworth band-pass filter is used for temporal filtering in the motion prompt generation, and the cut-off frequency for motion magnification is set between 0.4 and 4 Hz, corresponding to heart rate per minute. Additionally, the leave-one-subject-out (LOSO) protocol is utilized for evaluation, whereby one subject is reserved for validation while the remaining subjects are used for training.
\begin{figure*}[t]
    \centering
    \includegraphics[width=0.99\textwidth]{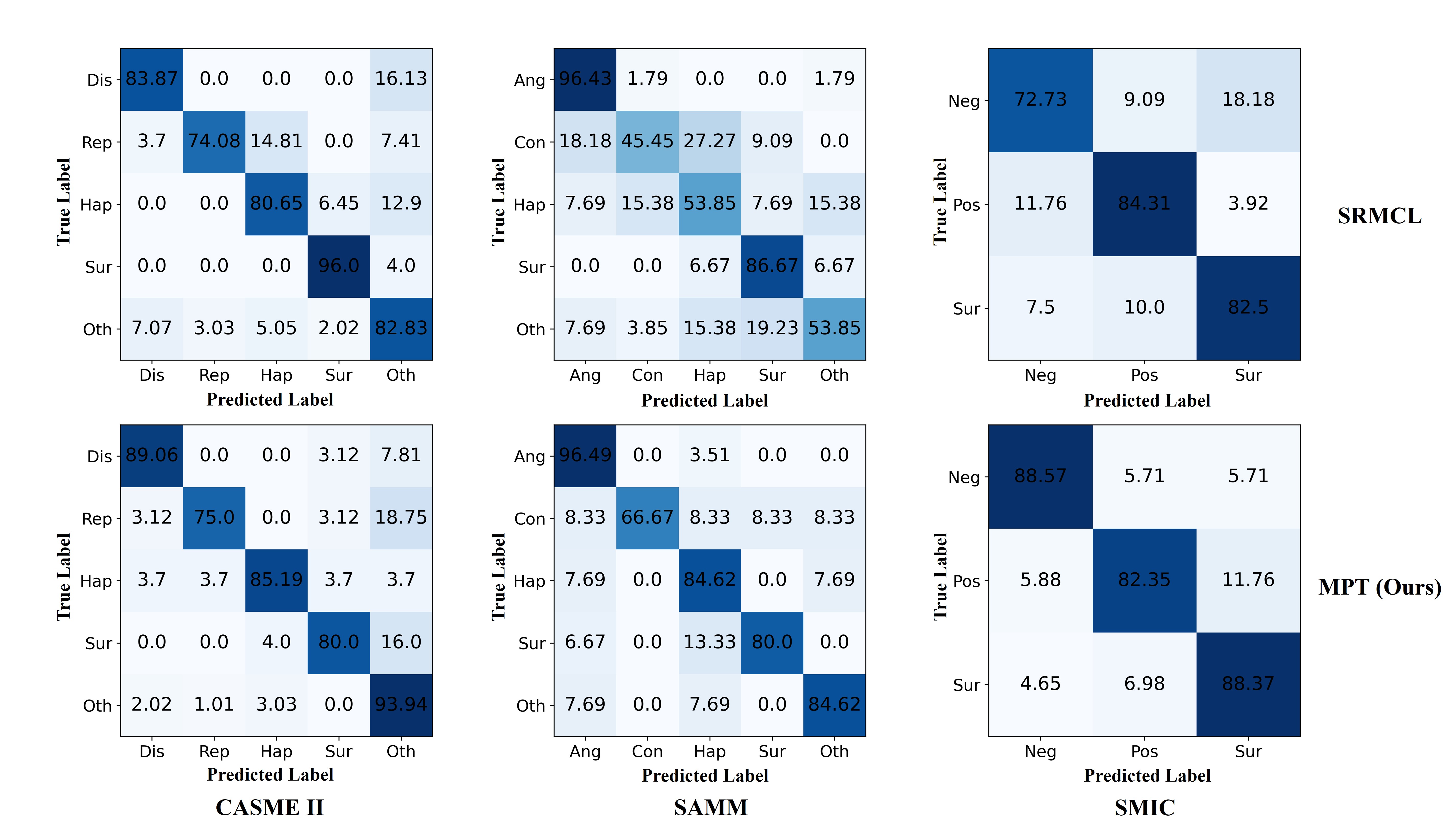}
    \caption{
      Confusion matrices of vision transformer-based SRMCL~\cite{bao2024boosting} and the proposed MPT on CASME II, SAMM, and SMIC databases.
    }
    \label{fig_cm}
\end{figure*}

\begin{table*}[ht]
\caption{Comparison results on recognition accuracy (\%)/F1-score (\%) between the proposed MPT and other transfer learning-based methods.}
\centering
\begin{tabular}{@{}c|c|ccc@{}}
\toprule
Method                                 & Year & CASME II      & SAMM           & SMIC          \\ \midrule
KD~\cite{sun2020dynamic}               & 2020 & 72.61 / 67.00 & -              & 76.06 / 71.00 \\
MicroNet~\cite{xia2020learning}        & 2020 & 75.60 / 70.10 & 74.10 / 73.60  & 76.80 / 74.40 \\
M2MTNet~\cite{xia2021micro}            & 2021 & 79.10 / 74.80 & 76.70 / 76.40  & 78.60 / 77.80 \\
NSPT~\cite{lee2022n}                   & 2022 & -             & -              & 79.47 / 79.37 \\
PLMaM-Net~\cite{wang2024progressively} & 2024 & 82.52 / 79.98 & 76.47 / 69.61  & 82.32 / 82.21 \\ \midrule
\textbf{MPT (Ours)}                    & 2024 & \textbf{87.85 / 86.04} & \textbf{87.50 / 85.04}  & \textbf{86.59 / 86.01} \\ \bottomrule
\end{tabular}
\label{tb_transcompare}
\end{table*}
Two common evaluation metrics are involved in the experiments, i.e., the recognition accuracy (Acc) and F1-score.
To be specific, assume that $TP_{c}$, $FP_{c}$, and $FN_{c}$ denote the true positive, false positive, and false negative rates for the $c$-th class,
the recognition accuracy and F1-score can be calculated as: 
\begin{equation}
 A c c=\frac{\sum_{c=1}^{C} T P_{c}}{N},
\end{equation}
\begin{equation}
 F 1-score=\frac{1}{C} \sum_{c=1}^{C} \frac{2 P_{c} \times R_{c}}{P_{c}+R_{c}},
\end{equation}
in which
\begin{equation}
P_{c}=\frac{T P_{c}}{T P_{c}+F P_{c}},
\end{equation}
\begin{equation}
R_{c}=\frac{T P_{c}}{T P_{c}+F N_{c}},
\end{equation}
where $N$ is the total number of samples of all subjects, $C$ is the number of categories in the datasets. Besides these, we also calculate the tunable parameters (\#Tparas) and GFLOPs during some comparison experiments to compare the model complexity and efficiency.

\subsection{Comparison with State-of-the-art Methods}
In this section, we conduct a series of comparison experiments and list the results to evaluate the proposed MPT.
\\
\textbf{Recognition results comparison.}
We present the results achieved in recognition accuracy (Acc) and mean F1-score 
under the leave-one-subject-out (LOSO) protocol on the CASME II, SAMM, and SMIC databases, 
which can be seen in Table~\ref{tb_casme}, Table~\ref{tb_samm} and Table~\ref{tb_smic}, respectively.
Meanwhile, a lot of state-of-the-art methods are chosen for comparison, including handcrafted methods (LBP-TOP~\cite{LBP-TOP-SAMM} LBP-SIP~\cite{LBP-SIP}, STCLQP~\cite{STCLQP}) as well as recent CNN and Transformer-based deep methods (\textit{e.g.}, 
 Alexnet~\cite{AlexNet}, 
 AUGCN~\cite{lei2021micro},
 Knowledge Distillation (KD)~\cite{sun2020dynamic}, DSSN~\cite{DSSN},
 TSCNN~\cite{ThreeStream}, Graph-TCN~\cite{GraphTCN}, SLSTT~\cite{SLSTT}, CMNET~\cite{wei2023cmnet}, LTR3O~\cite{zhu2025learning}, $\mu$-BERT~\cite{nguyen2023micron}, PLMaM-Net~\cite{wang2024progressively} and SRMCL~\cite{bao2024boosting}).

According to the results, the proposed MPT achieves the best performance in all cases. 
Specifically, on the CASME II database, our MPT outperforms the previous state-of-the-art method FFDIN by a large margin of 2.08\%/2.95\% in Acc and F1-score. On the SAMM database, our method gains improvements of 7.35\% and 9.30\% in Acc and F1-score than the state-of-the-art LTR3O, respectively. Our MPT also achieves the best performance in the SMIC-HS database and outperforms the state-of-the-art PLMaM-Net~\cite{wang2024progressively} for 4.27\% and 4.69\% in Acc and F1-score. 
The reasons for these superior results can be attributed to several factors. On the one hand, our MPT leverages pre-trained knowledge of LMs and proposes motion prompts to enhance subtle motion recognition, the strong power of LMs successfully improves the MER performance. 
On the other hand, a group adapter is presented in our MPT to alleviate domain gaps between LMs and MEs further, which can handle information from different domains efficiently. Furthermore, in contrast to the previous methods, the proposed MPT uses a small number of tunable parameters that can prevent the data-limited MER model from being trapped in overfitting while preserving the rich knowledge of LMs, thus it can achieve more effective performance.
\\
\textbf{Confusion matrices.}
We especially show the confusion matrices of the state-of-the-art SRMCL~\cite{bao2024boosting} method and our MPT in Fig~\ref{fig_cm}, both of which are built on the vision transformer structure~\cite{vit}. As the figure shows, it is clear that the proposed MPT model generally outperforms the SRMCL method across all datasets and categories. For example, MPT gains a large margin on the negative class on SMIC.
According to the results, we can observe several phenomena.
The introduction of motion cues and adapters for prompt tuning in the MPT model significantly enhances its performance over the SRMCL method across multiple ME recognition datasets. While both methods leverage the powerful ViT architecture, the addition of motion cues allows MPT to capture dynamic aspects of MEs, and the group adapter facilitates more effective prompt tuning. This results in more robust and accurate classification, as evidenced by the higher accuracy rates and reduced confusion in categories that are challenging for SRMCL. 
The improved performance of MPT across CASME II, SAMM, and SMIC datasets demonstrates its superior ability to effectively adapt LMs to accurately recognize MEs. That is to say, the proposed fine-tune strategy in MPT, including motion prompt generation and the group adapter, which leverages the powerful ability of pretrained LMs and adapts them to MER, is much better than those that directly use the vision transformer as a backbone to train it from scratch.
\begin{table}[]
\centering
\caption{Efficiency comparison with other state-of-the-art methods. \#Tparas denotes the number of tunable parameters.}
\begin{tabular}{@{}c|c|c|c|c@{}}
\toprule
Methods    & \#Tparas (M) & GFLOPs & Acc (\%) & F1 (\%) \\ \midrule
OFF-Apexnet~\cite{OFF-ApexNet} & 2.7         & 0.1   & 74.60    & 71.04   \\
AMAN~\cite{wei2022novel}       & 8.4          & 81.5 & 79.87    & 77.08   \\
VGG-16~\cite{VGG-16}    & 138.4       & 31.0  & -      & 59.64   \\
FeaRef~\cite{zhou2022feature}     & 26.0        & 0.1   & -      & 73.72   \\
SLSTT~\cite{SLSTT}      & 93.1        & 791.4 & 75.00    & 74.00   \\
SRMCL~\cite{bao2024boosting}      & 315.7        & 124.3 & 78.98    & 78.87   \\ \midrule
MPT        & 5.5         & 18.7 & 86.59    & 86.01   \\ \bottomrule
\end{tabular}
\label{tb_eff}
\end{table}
\begin{table}[ht]
    \centering
    \caption{Effects of different learning strategies on the SMIC database. \#Tparas denotes the number of tunable parameters.}
    \begin{tabular}{@{}c|ccccc@{}}
    \toprule
     Method             & \#Tparas (M)  & Acc($\%$) & F1-Score (\%) \\ \midrule
    w/o pretraining     & 86.4          & 73.17     & 72.35    \\
    full fine-tuning    & 86.4          & 77.44     & 76.65    \\
    partial fine-tuning    & 0.003         & 71.95     & 70.27    \\
    motion prompt       & 0.2           & 82.92     & 82.33    \\
    group adapter       & 3.5           & 79.27     & 78.43    \\
    MPT (Ours)          & 5.5           & 86.59     & 86.01    \\ \bottomrule
    \end{tabular}
    \label{tb_abl}
\end{table}
\begin{table}[ht]
    \centering
    \caption{Effects of different prompts on the SMIC database. \#Tparas denotes the number of tunable parameters.}
    \begin{tabular}{@{}c|ccccc@{}}
    \toprule
    Type           &\#Tparas (M)  & Acc($\%$) & F1-Score (\%)  \\ \midrule
    w/o prompt     & 3.5          & 79.27     & 78.43    \\
    VPT-Shallow    & 5.5          & 81.10     & 80.84    \\
    VPT-Deep       & 5.8          & 82.32     & 81.79    \\ 
    MPT (Ours)     & 5.5          & 86.59     & 86.01    \\ \bottomrule
    \end{tabular}
    \label{tb_ablprompt}
\end{table}
\begin{table}[ht]
    \centering
    \caption{Effects of different adapters in MTN on the SMIC database. \#Tparas denotes the number of tunable parameters.}
    \begin{tabular}{@{}c|ccccc@{}}
    \toprule
    Type                &\#Tparas (M) & Acc($\%$) & F1-Score (\%) \\ \midrule
    MPT (w/o adapter)   & 0.2         & 82.92     & 82.33   \\
    primitive adapter   & 2.0         & 84.15     & 83.56   \\
    MPT (Ours)          & 5.5         & 86.59     & 86.01   \\ \bottomrule
    \end{tabular}
    \label{tb_abladapter}
\end{table}
\\
\textbf{Comparison with other transfer learning-based MER methods.}
Several transfer learning-based MER methods are listed for comparison, including KD~\cite{sun2020dynamic}, MicroNet~\cite{xia2020learning}, M2MTNet~\cite{xia2021micro}, NSPT~\cite{lee2022n}, and PLMaM-Net~\cite{wang2024progressively}. As shown in Table~\ref{tb_transcompare}, the proposed MPT outperforms all the transfer learning-based methods. The main reason is that all the existing transfer learning methods just construct similar network structures to mimic the facial appearance and texture representation of macro-expressions, which will inevitably fall into overfitting due to the limited ME data number thus they can not narrow the gap between MaE and ME to obtain ME features. However, our MPT uses a parameter-efficient way to fine-tune the LMs, making them adapt to the MER domain, which utilizes the rich information in LMs without knowledge loss and copes with the issue of limited data number of ME successfully.
\\
\textbf{Efficiency comparison.}
Table~\ref{tb_eff} provides a comprehensive evaluation of recent SOTA models based on the number of tunable parameters (\#Tparas), computational complexity (GFLOPs), accuracy (Acc), and F1-score. 
In terms of the results, MPT achieves the highest accuracy of 86.59\% and F1 score of 86.01\%  with only 5.5 M tunable parameters and 18.7 GFLOPs, showing its parameter efficiency. This performance is excellent given its moderate complexity, as MPT outperforms many large models, including VGG-16 (with 138.4 M tunable parameters) and SRMCL (with 315.7 M parameters), both of which have much lower accuracy results. In a word, MPT demonstrates a superior balance between computational demand and performance compared to other methods trained from scratch, which proves the effectiveness of the presented strategies in MPT, as it successfully utilizes the powerful ability of LMs and adapts them to the MER domain to handle the limited data issue efficiently.
\subsection{Ablation Study}
Extensive experiments are conducted to make a more in-depth analysis of the proposed MPT.
\\
\textbf{Effects of different learning strategies.} In Table~\ref{tb_abl}, we investigate different learning strategies. In the first row, we select a ViT-B/16 model~\cite{vit} as the backbone without any pretraining. 
Compared with this setting, the ViT-B/16 pre-trained on ImageNet-21K~\cite{russakovsky2015imagenet} and fully fine-tuned on SMIC (the second row) achieves improvements of 4.27\% in Acc and 4.30\% in F1-Score, which shows that the rich knowledge in LMs does benefit the MER issue. 
However, in the third row, we only fine-tune the classification head rather than the whole model, declining the performance over the baseline method. A reason is that MER datasets only contain limited data, and thus, large models are easy to overfit. Meanwhile, some pretrained knowledge is forgotten during full fine-tuning, while partial fine-tuning can not adapt to the MER domain well. In the fourth row, we add our motion prompts to the partly fine-tuned model, which yields very large improvements in both Acc and F1-Score. Compared with the partly fine-tuned model, our group adapter also achieves gains of 7.32\% in Acc. and 8.16\% in F1-Score. These demonstrate the effectiveness of our motion prompts and group adapter. Finally, our MPT with both motion prompts and the group adapter shows the best performance, which outperforms the partly fine-tuned model with important improvements by 14.64\% in Acc and 15.74\% in F1-Score. In a word, both the presented motion prompts and group adapters can improve the ME feature learning a lot.
\\
\textbf{Effects of different prompts.} 
We compare the proposed MPT with the classic visual prompt methods VPT-Shallow and VPT-Deep~\cite{vpt}, which insert randomly initialized prompts to the LMs. The results can be seen in Table~\ref{tb_ablprompt}, and we can get several conclusions. Firstly, all prompt-based methods outperform the model without prompts, which proves that using prompt tuning can effectively adapt LMs to MER. Secondly, the proposed MPT significantly improves performance compared to the two classic visual prompt tuning methods. The main reason is that since the common visual prompts used in VPT are just randomly initialized, it can not model the subtle motion of MEs and capture the robust ME representation well. However, our MPT introduces parameter-limited motion cues as visual prompts for LMs and gains an important improvement over the other methods, which shows it can effectively model the subtle motion of MEs.
\\
\textbf{Effects of different adapters.} We compare the proposed group adapter with the primitive adapter~\cite{vpt} and MPT without the adapter.
According to the results in Table~\ref{tb_abladapter}, the addition of adapters improves the performance, and the proposed group adapter outperforms the primitive one by 3.67\% and 3.68\% in Acc and F1-score, respectively. The reason is that our group adapter more specifically adapts each type of feature since it utilizes the group structure to process different types of information in the transformer layer effectively.
\begin{figure}[t]
    \centering
    \includegraphics[width=0.48\textwidth]{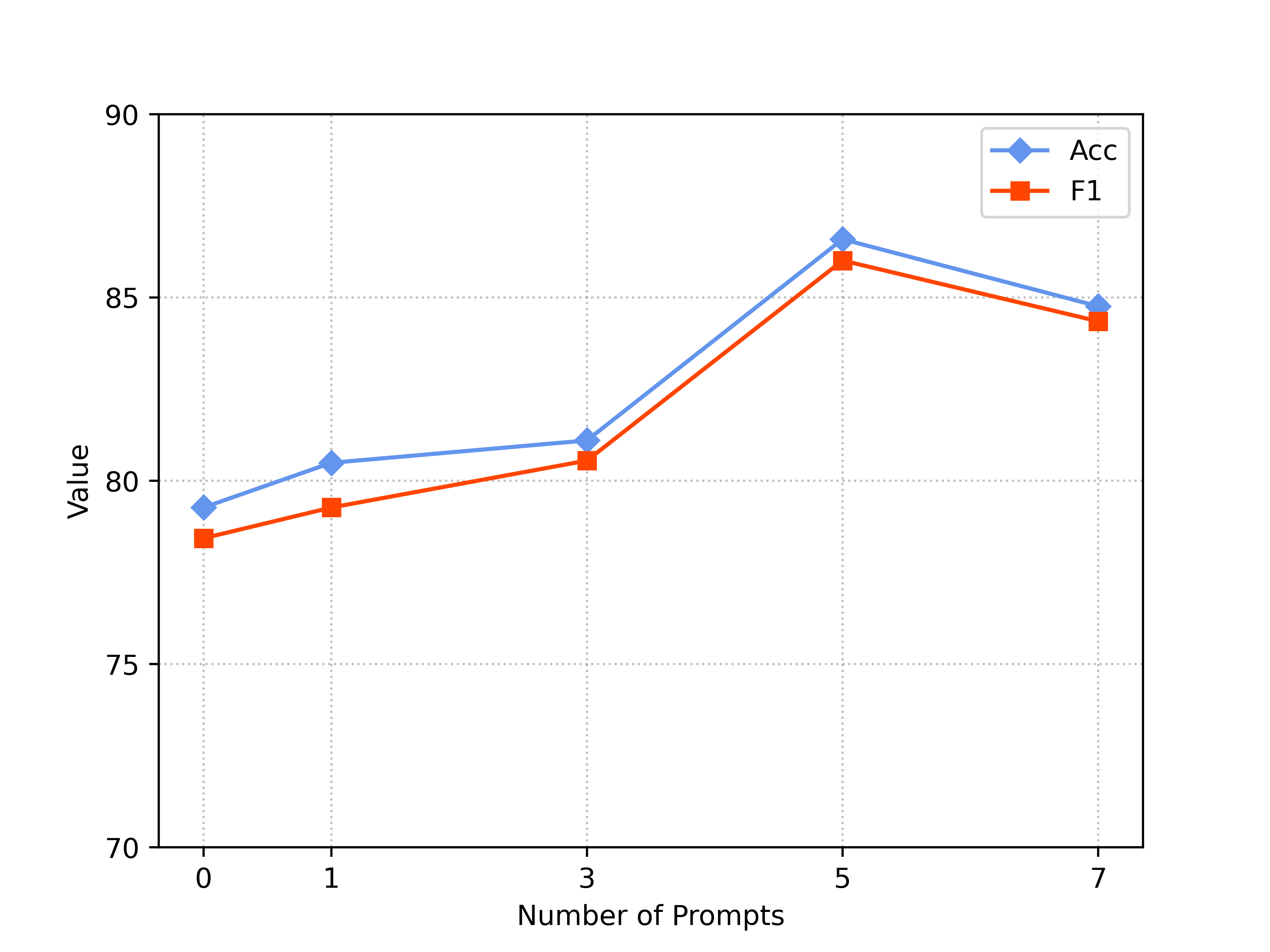}
    \caption{Recognition accuracy (Acc) and F1-score achieved by different numbers of motion prompts on the SMIC database. The best results is achieved when the number of prompts is set to 5.}
    \label{fig_albnumprompt}
\end{figure}
\begin{figure*}[t]
    \centering
    \includegraphics[width=0.98\textwidth]{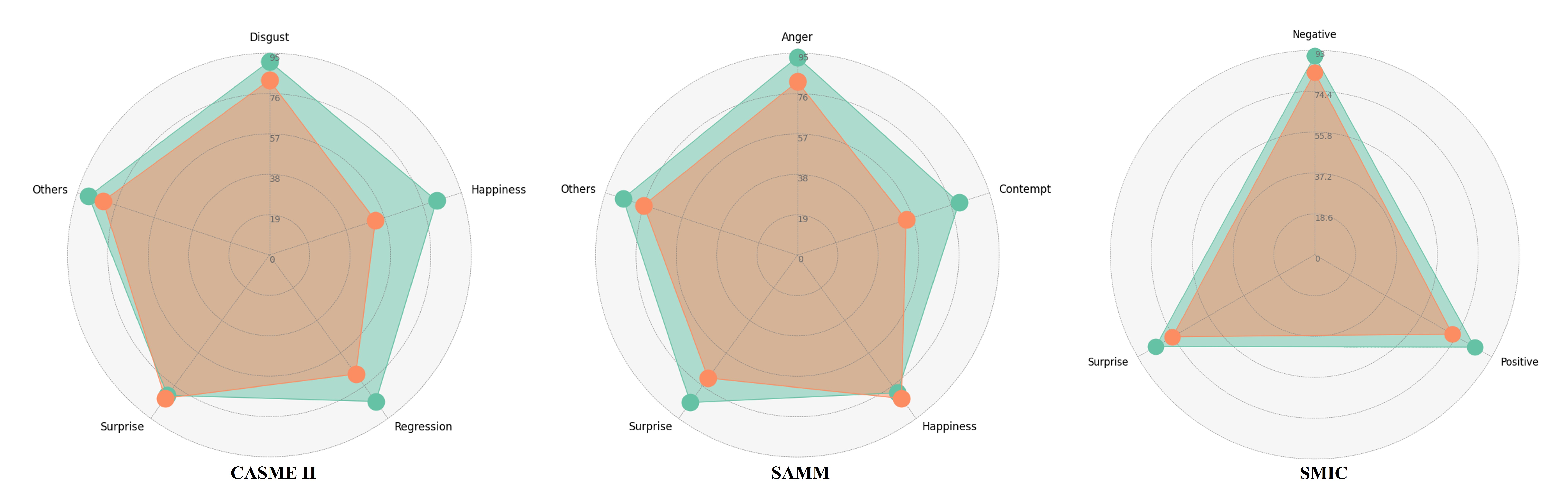}
    \caption{Comparison of the proposed MPT with baseline at class level. The F1-score results are visualized on CASME II, SAMM, and SMIC databases, respectively.}
    \label{fig_radar}
\end{figure*}
\begin{table}[]
    \centering
    \caption{Ablation study of different pre-trained models on the SMIC database.}
    \begin{tabular}{@{}c|cccc@{}}
    \toprule
    Model          & Acc($\%$)  & F1-Score (\%) \\ \midrule
    ViT-B/16       & 86.59      & 86.01         \\
    ViT-B/32       & 84.76      & 83.95         \\ 
    CLIP-ViT-B/16  & 85.37      & 84.64         \\ 
    CLIP-ViT-B/32  & 83.54      & 82.56     \\ \bottomrule
    \end{tabular}
    \label{tb_ablpretrain}
\end{table}
\\
\textbf{Effects of the number of motion prompts.} 
We compare different numbers of our motion prompts, and the results are shown in Fig.~\ref {fig_albnumprompt}. Models with 1, 3, 5, and 7 prompts outperform the model without motion prompts. Both accuracy and F1-score show an increasing trend as the number of prompts increases from 0 to 5. This indicates that the inclusion of more prompts generally enhances the model's performance. After 5 prompts, there is a slight decrease in both metrics when the number of prompts is increased to 7. This suggests that beyond a certain point, adding more prompts may not continue to improve the model's performance and could potentially introduce redundant motion information or noise. Anymore, the model with 5 motion prompts achieves the best performance with 86.59\%/86.01\% on the Acc/F1-score.
\\
\textbf{Effects of different LMs.} 
As shown in Table~\ref{tb_ablpretrain}, two types of ViT are utilized, including ViT-B/16 and ViT-B/32 trained on the ImageNet21K~\cite{vit} and CLIP~\cite{clip}. ViT-16-based models achieve better performance than ViT-B/32-based ones since smaller patches retain more fine-grained details of the facial regions, which can be crucial for recognizing subtle features and textures that contribute to the MER task, while the larger patches might miss these details, leading to lower performance.
Meanwhile, the results of CLIP-based models are lower than those pretrained on ImageNet21K. 
Furthermore, most of the cases outperform the state-of-the-art methods on the SMIC database shown in Table~\ref{tb_smic}. This can be the result of the MER issue with a stable set of classes, while the common pretrained ViT-B/16 is likely to be more beneficial since it may contain more visual information like texture and appearances. This fine-tuning can help the model learn to detect the subtle features specific to MEs.
\begin{table}[t]
\centering
\caption{Effects of different magnification factors on the SMIC database.}
\begin{tabular}{@{}c|c|c@{}}
\toprule
Factor ($\beta$) & Acc (\%) & F1 (\%) \\ \midrule
0     & 81.10    & 80.54  \\
10    & 85.37    & 84.79   \\
20    & 86.59    & 86.01   \\
30    & 81.70    & 81.44   \\
50    & 78.05    & 77.56   \\ \bottomrule
\end{tabular}
\label{tb_magfactor}
\end{table}
\begin{table}[t]
\centering
\caption{Effects of different reduction factors on the SMIC database.}
\begin{tabular}{@{}c|c|c@{}}
\toprule
Factor ($\eta$) & Acc (\%) & F1 (\%) \\ \midrule
4     & 85.98    & 85.32  \\
8     & 86.59    & 86.01   \\
12    & 84.76    & 84.05   \\ \bottomrule 
\end{tabular}
\label{tb_eta}
\end{table}
\\
\textbf{Effects of hyper-parameters.}
Table~\ref{tb_magfactor} illustrates the influence of varying magnification factors (i.e., $\beta$) of both Acc and F1-score on the SMIC database. The results prove the effectiveness of motion magnification and reveal a clear trend for determining the most suitable factor $\beta$. At $\beta$ is set to 10, MPT achieves an accuracy of 85.37\% and an F1-score of 84.79\% with a large margin to the experiment without motion magnification ($\beta = 0$), which only achieves 81.10\%/80.54\% in Acc/F1-score. Setting the factor to 20 further improves the results, with the best Acc and F1-score of 86.59\% and 86.01\%, respectively. However, further increases in $\beta$ would result in a severe decline in performance, with $\beta$ set to 30 and 50 reducing Acc to 81.70\% and 78.05\%, and F1-scores to 81.44\% and 77.56\%, respectively. The phenomenon suggests that proper magnification enhances discriminative motion feature representation, and an overly large $\beta$ would cause the information of the original ME sequence (like facial textures and appearances) to be damaged.

Table~\ref{tb_eta} presents the result of the reduction factor $\eta$ in the proposed group adapters. When set to 8, MPT achieves the highest accuracy of 86.59\% and F1 score of 86.01\%, suggesting this configuration can effectively reduce computational complexity while maintaining excellent performance. Conversely, factors of 4 and 12 result in lower results. It indicates that an appropriate $\eta$ can strike a balance between performance and efficiency.
\begin{figure}[ht]
    \centering
    \includegraphics[height=0.38\textwidth,width=0.45\textwidth]{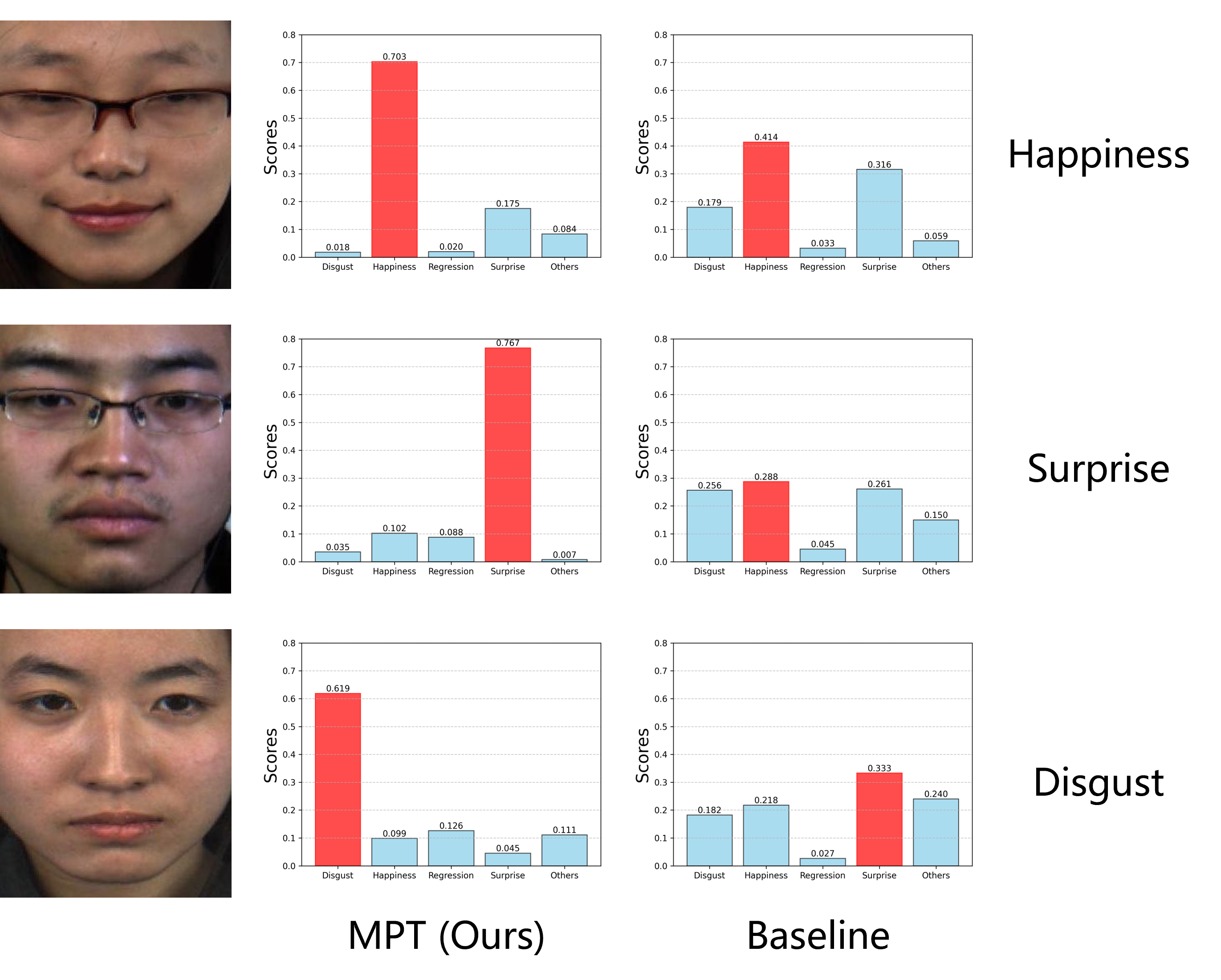}
    \caption{
      Comparison of the prediction cases of the proposed MPT and baseline method. From top to bottom, there are samples selected from CASME II~\cite{CASMEII} related to happiness, surprise, and disgust, respectively.
    }
    \label{prediction}
\end{figure}
\subsection{Visualization Analysis}
In this section, we conduct a series of visualization experiments to further evaluate the proposed modules in MPT.
\\
\textbf{Prediction results visualization.}
To further evaluate the effectiveness of the proposed modules, the overall F1-score and detailed breakdown of class accuracy of the proposed MPT and baseline on each emotion type on CASME II, SAMM, and SMIC-HS databases are visualized in Fig.~\ref{fig_radar}. 
Specifically, the model fine-tunes the pretrained primitive ViT-B/16 without inserting the proposed motion prompt, and the group adapter is selected as a baseline comparison.
In terms of the results, we can demonstrate the performance of the proposed MPT at the class level, and it significantly improves the performance in the confusable regression type in the CASME II database and the contempt type in the SAMM database. That is to say, the proposed modules successfully represent the subtle motions in these types of MEs.
Furthermore, several samples from the CASME II database predicted by the proposed MPT and baseline method are shown in Fig.~\ref{prediction}.
According to the figure, we can find that the proposed MPT achieves a higher prediction score on the distinguishable happiness sample and also predicts the right results on the surprise and disgust samples,
while the baseline method obtains the wrong predictions on the confusable surprise and disgust samples.
Hence, the proposed MPT is more robust since it introduces additional distinguishable facial movement information to the LMs that would benefit from representing subtle MEs.
\begin{figure}[t]
    \centering
    \includegraphics[width=0.38\textwidth]{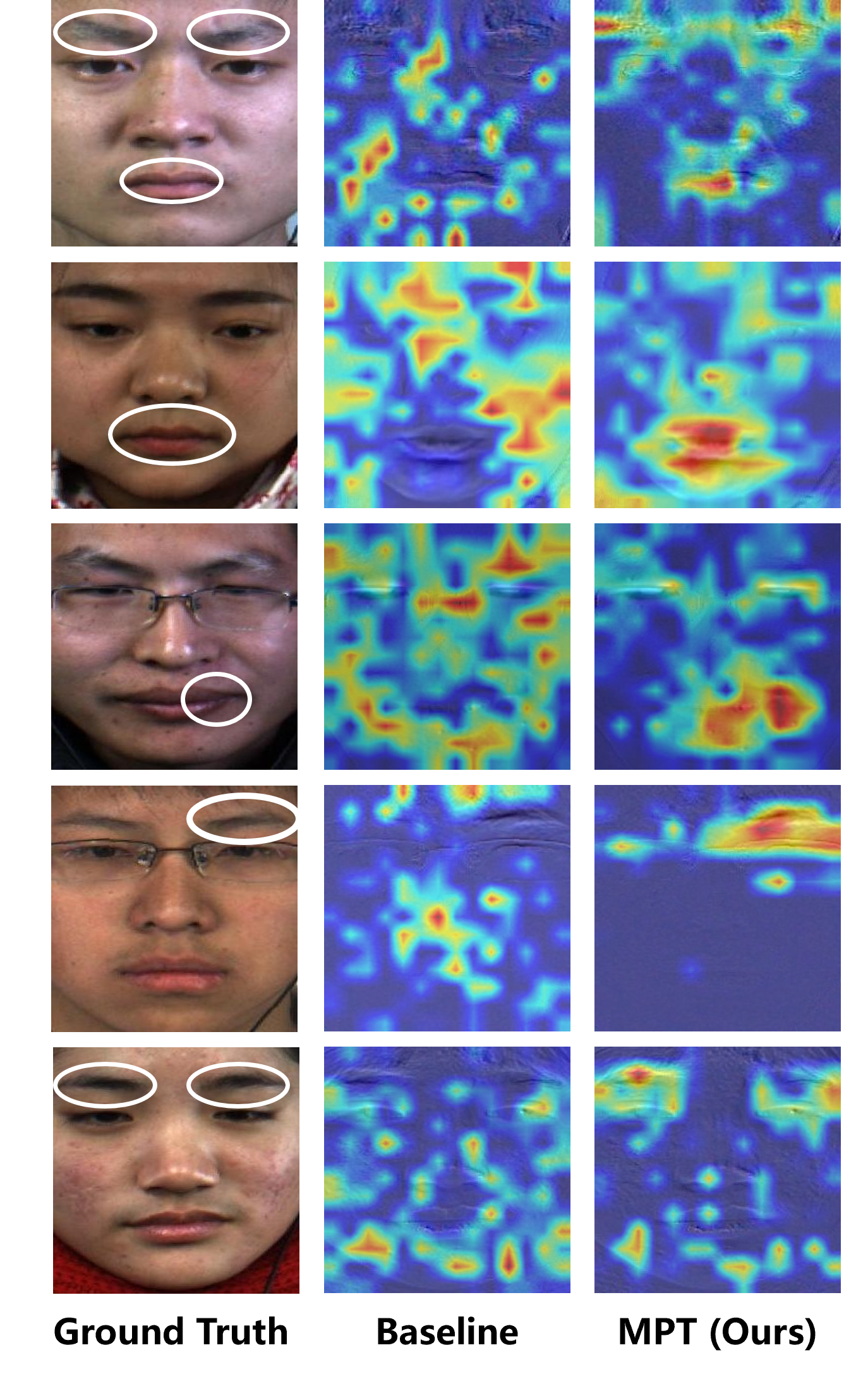}
    \caption{Visualization results of feature heatmaps of several samples in the CASME II dataset. Each row denotes an input and the corresponding activation maps. From left to right, we show the ground truth samples of AU ground truth, the results obtained by the baseline method, and the results obtained by the proposed MPT, respectively.}
    \label{fig_cam}
\end{figure}
\begin{figure}[ht]
    \centering
      \includegraphics[width=0.48\textwidth]{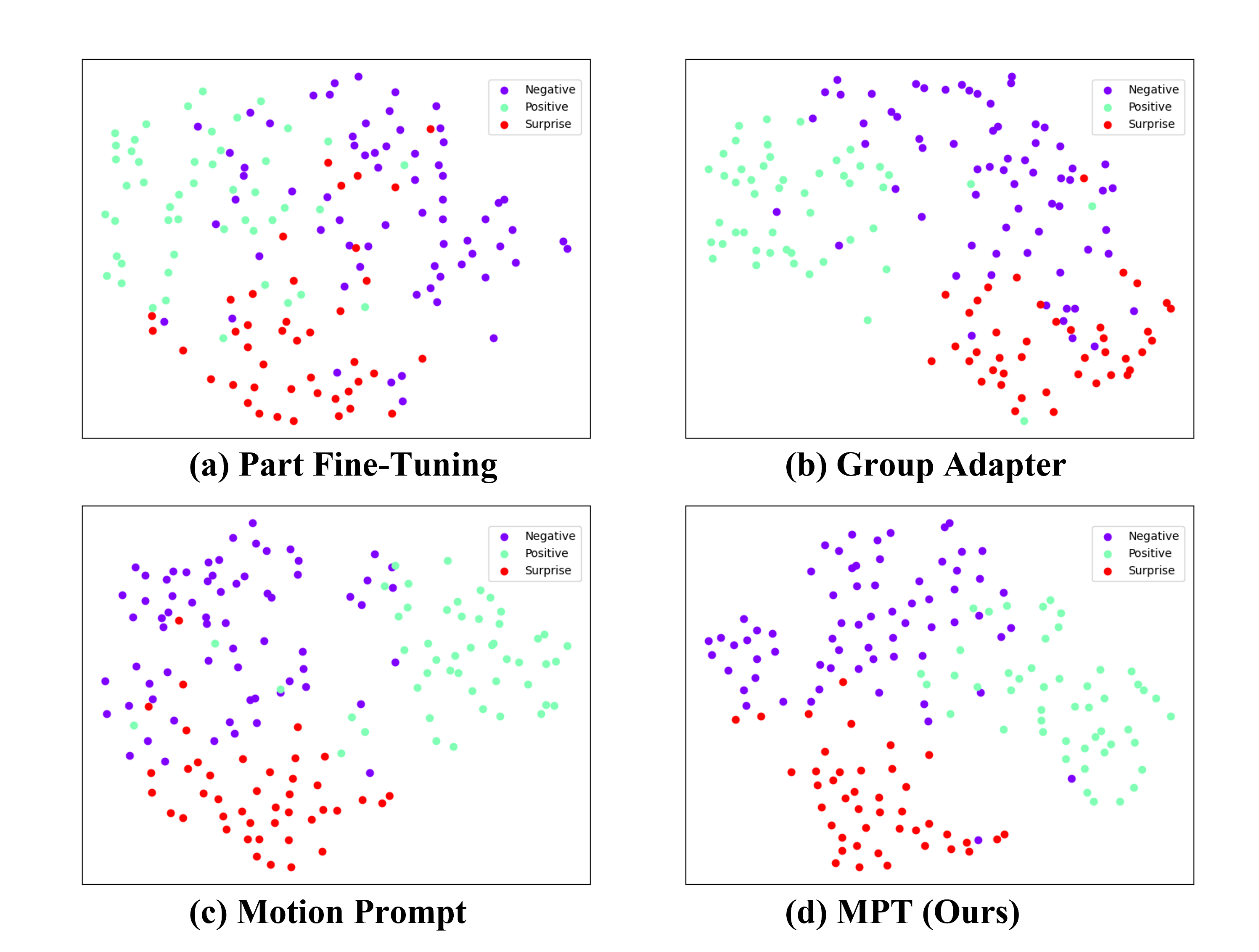}
    \caption{t-SNE visualization comparison results on the SMIC-HS database. Fig. (a) denotes the experiment that does part fine-tuning on a pretrained ViT-B/16, while Fig. (b), (c), (d) denote the experiments that conduct fine-tuning on the pretrained ViT-B/16 with the proposed group adapter, motion prompt and the whole MPT, respectively. }
    \label{tsne}
\end{figure}
\\
\textbf{Visualization results of feature heatmaps.}
To better evaluate the learned features, we visualize the feature heatmaps using GradCAM~\cite{gradcam}.
As shown in Fig~\ref{fig_cam}, we select the common ViT~\cite{vit} as a baseline and compare its activation heatmap visualization results with our MPT.
It can be seen from the results that the proposed MPT focuses more on the AU region relating to MEs than the baseline method.
For example, in the first row, the activation map of the proposed MPT focuses more on the brows and lips, which is consistent with the provided AU ground truth (Brow Lowerer and Upper Lip Raiser). In the second row,
the region near the lips is highly activated, which is consistent with the AU 25 (Lips Part) annotation.
The results prove that our proposed MPT has a stronger ability to capture the subtle motion cues in MEs than the common vit structure since it introduces magnified motion information as prompts to adapt the LM to learn the ME feature, and the introduced motion prompts are effectively absorbed in the structure of the LM through the light-weight adapter layers. In contrast, the common ViT can not represent the subtle motion of ME well since it is trained on the RGB image directly without introducing extra motion cues.
\\
\textbf{Feature distribution.}
To evaluate the effectiveness of the proposed MPT, we apply t-SNE~\cite{tsne} to visualize the feature embedding under different settings as shown in Fig~\ref{tsne}. 
To be specific, a pretrained ViT-B/16 is utilized as the backbone.
Fig. (a) denotes the experiment that does part fine-tuning on the classification head, while Fig. (b), (c), (d) denote the experiments that conduct fine-tuning on the pretrained ViT-B/16 with the proposed group adapter, motion prompt, and the whole MPT, respectively.
According to the results, both motion prompt and group adapter extract more distinguishable feature boundaries with three class clusters.
Feature embedding of part fine-tuning on serious overlap among the three classes, while the confusable surprise and positive samples are separated clearly in Fig.~\ref{tsne}(c) and (d), which means the generation of motion prompts plays a significant role in assisting the LMs to capture the subtle motion cues.
Furthermore, the proposed MPT can obtain a clearer class boundary
which demonstrates that the proposed MPT can learn more representative characteristics of the MER issue.

\section{Conclusion}
In this paper, we have proposed a novel motion prompt tuning network model (MPT) for adapting large pre-training models (LMs) to the data-limited MER problem. Based on a frozen LM,
MPT first utilizes a motion prompt generation module to magnify subtle motions and then generates motion prompts by aggregating the temporal salient snapshots.
Furthermore, a simple yet effective group adapter (GA) is presented to reduce the gaps between pretrained LMs and MER data efficiently.
Extensive experiments conducted on several databases show that the proposed MPT outperforms state-of-the-art MER methods, which demonstrates that the proposed motion prompt generation and tuning methods can help the LMs focus on subtle motion cues and enhance the MER performance effectively and efficiently.

\newpage
\bibliographystyle{IEEEtran}
\bibliography{IEEEtran}

\end{document}